\documentclass[10pt,twocolumn,letterpaper]{article}

\usepackage{cvpr}              
\usepackage{multirow} 
\usepackage{marvosym}
\definecolor{cvprblue}{rgb}{0.21,0.49,0.74}
\usepackage[pagebackref,breaklinks,colorlinks,allcolors=cvprblue]{hyperref}
\makeatletter
\newcommand\blfootnote[1]{%
  \begingroup
  \renewcommand\thefootnote{}\footnote{#1}%
  \addtocounter{footnote}{-1}%
  \endgroup
}
\makeatother
\def\paperID{} 
\def\confName{CVPR}
\def\confYear{2026}

\title{\textcolor{red}{L}\textcolor{green}{o}\textcolor{blue}{L}\textcolor{orange}{A}: \textcolor{red}{L}ong H\textcolor{green}{o}rizon \textcolor{blue}{L}atent \textcolor{orange}{A}ction Learning for General Robot Manipulation}

\author{Xiaofan Wang$^{1,2\dagger}$ \quad \text{Xingyu Gao}$^{1,2}$\textsuperscript{\Letter}  \quad \text{Jianlong Fu}$^{3 }$ \quad \text{Zuolei Li}$^{1,2\dagger}$ \\ \text{Dean Fortier}$^{3}$ \quad \text{Galen Mullins}$^{3}$ \quad \text{Andrey Kolobov}$^{3}$ \quad \text{Baining Guo}$^{3}$\\[3pt]
$^1$~\text{Institute of Microelectronics, Chinese Academy of Sciences} \\ $^2$~\text{University of Chinese Academy of Sciences} \quad
$^3$~\text{Microsoft Research}  \\
{\tt\small \{wangxiaofan24, lizuolei24\}@ime.ac.cn, gxy9910@gmail.com } \\
{\tt\small \{jianf, v-defortier, galenmullins, akolobov, bainguo\}@microsoft.com}}

\begin{document}
\maketitle

\blfootnote{$\dagger$ Work conducted during internship at Microsoft Research}
\blfootnote{\textsuperscript{\text{\Letter}} Corresponding author: gxy9910@gmail.com}

\begin{abstract}

The capability of performing long-horizon, language-guided robotic manipulation tasks critically relies on leveraging historical information and generating coherent action sequences. However, such capabilities are often overlooked by existing Vision-Language-Action (VLA) models. To solve this challenge, we propose \textbf{LoLA} (\textbf{L}ong H\textbf{o}rizon \textbf{L}atent \textbf{A}ction Learning), a framework designed for robot manipulation that integrates long-term multi-view observations and robot proprioception to enable multi-step reasoning and action generation. We first employ Vision-Language Models to encode rich contextual features from historical sequences and multi-view observations. We further introduces a key module, \textbf{State-Aware Latent Re-representation}, which transforms visual inputs and language commands into actionable robot motion space. Unlike existing VLA approaches that merely concatenate robot proprioception (e.g., joint angles) with VL embeddings, this module leverages such robot states to explicitly ground VL representations in physical scale through a learnable ``embodiment-anchored" latent space.
We trained LoLA on diverse robotic pre-training datasets and conducted extensive evaluations on simulation benchmarks (SIMPLER and LIBERO), as well as two real-world tasks on Franka and Bi-Manual Aloha robots. Results show that LoLA significantly outperforms prior state-of-the-art methods (e.g., $\pi_0$), particularly in long-horizon manipulation tasks.

\end{abstract}    
\section{Introduction}
\label{sec:intro}
Recent advances in robotic manipulation have been largely driven by Vision-Language-Action (VLA) models \cite{liu2024rdt, zheng2024tracevla}, which leverage pre-trained Vision-Language Models (VLMs)  trained on vast web data \cite{yang2024qwen2, chen2024expanding, hu2024minicpm}. By exploiting the rich, multi-modal representations from VLMs, VLA models achieve strong cross-modal understanding, enabling them to interpret complex natural language instructions and ground them in visual observations \cite{vuong2023open, kim2024openvla}. This capability further allows impressive action generalization capabilities to novel objects, environments and tasks, surpassing the limitations of traditional task-specific robot policy models~\cite{chi2023diffusion}. Although promising results have been observed, most existing works focus on short-horizon tasks, which limits their applicability in real-world scenarios. Such tasks are typically completed in just a few seconds (e.g., pick up a blue cup) and therefore fail to capture complex, goal-oriented behaviors (e.g., pizza making). 

The key challenges in learning long-horizon tasks can be summarized as follows. First, long-horizon tasks require a coherent understanding of temporal context. Traditional single-frame models \cite{black2410pi0, li2024cogact, wen2024diffusion, wen2025tinyvla}, lacking this temporal awareness, struggle to accurately track multi-step states (e.g., pour water three times) or maintain action consistency over extended periods, often resulting in repetitive errors. Second, prolonged interactions between the robot and its environment drastically increase the likelihood of encountering unexpected, out-of-distribution situations \cite{xiao2019learning}. For instance, a minor perturbation early in the task (e.g., a cup being bumped) can accumulate over time, driving the robot into states far from the training distribution and ultimately causing policy failure. Third, long-horizon learning demands substantial resources for both training and roll-out. Collecting and testing minute-long data with frequent human resets are exponentially more expensive than short-horizon tasks, while processing long historical sequences during inference incurs heavy computational overhead.

To address these challenges, we propose LoLA (Long-horizon Latent Action Learning), a framework for long-horizon robot manipulation tasks. LoLA efficiently integrates long-term multi-view observations through high-fidelity Current Observation Encoding and downsampled Historical Motion Encoding, optimizing memory and computation while preserving essential historical context. To bridge the gap between vision-language perception and physical action spaces while mitigating the risk of robot state drift in the real world, we introduce a \textit{State-aware Latent Re-representation (SALR)} module that constructs a new latent space grounding pre-trained vision-language features to real robot actions. By anchoring this space to robot proprioception (e.g., joint angles and end-effector positions), \textit{SALR} explicitly aligns visual perception with real-world physical scales while constraining the learned representation to adhere to physically plausible motion patterns. This enables the model to efficiently leverage the physical-scale trajectory consistency in pre-training data to stabilize real-world inference. We further apply a \textit{Learnable Mask Operation} to actively suppress action-irrelevant noise (e.g., background distractors), and select the most critical action-relevant signals. These filtered features are then distilled into a compact representation, which serves as input to a downstream action expert for multi-step action generation. 

To validate LoLA, we pre-trained the proposed model on diverse, cross-embodiment robotic datasets \cite{o2024open, bu2025agibot}, and collected a novel long-horizon dataset of complex manipulation tasks, consisting of 28 real-robot tasks. These include 22 atomic sub-tasks organized into 7 sequential long-horizon tasks, and 6 additional continuous long-horizon tasks with an average duration of 2.3 minutes each. Results show that LoLA significantly outperforms state-of-the-art methods across diverse embodiments and environments, on both simulation benchmarks (SIMPLER \cite{li2024evaluating}, LIBERO \cite{liu2023libero}) and real-world robotic platforms (Franka Research 3 and Bi-Manual Aloha robots). It demonstrates strong robustness and generalization, particularly in long-horizon tasks, such as combined task sequences in the pizza-making process.

\begin{figure*}
  \centering
  \includegraphics[width=1\textwidth]{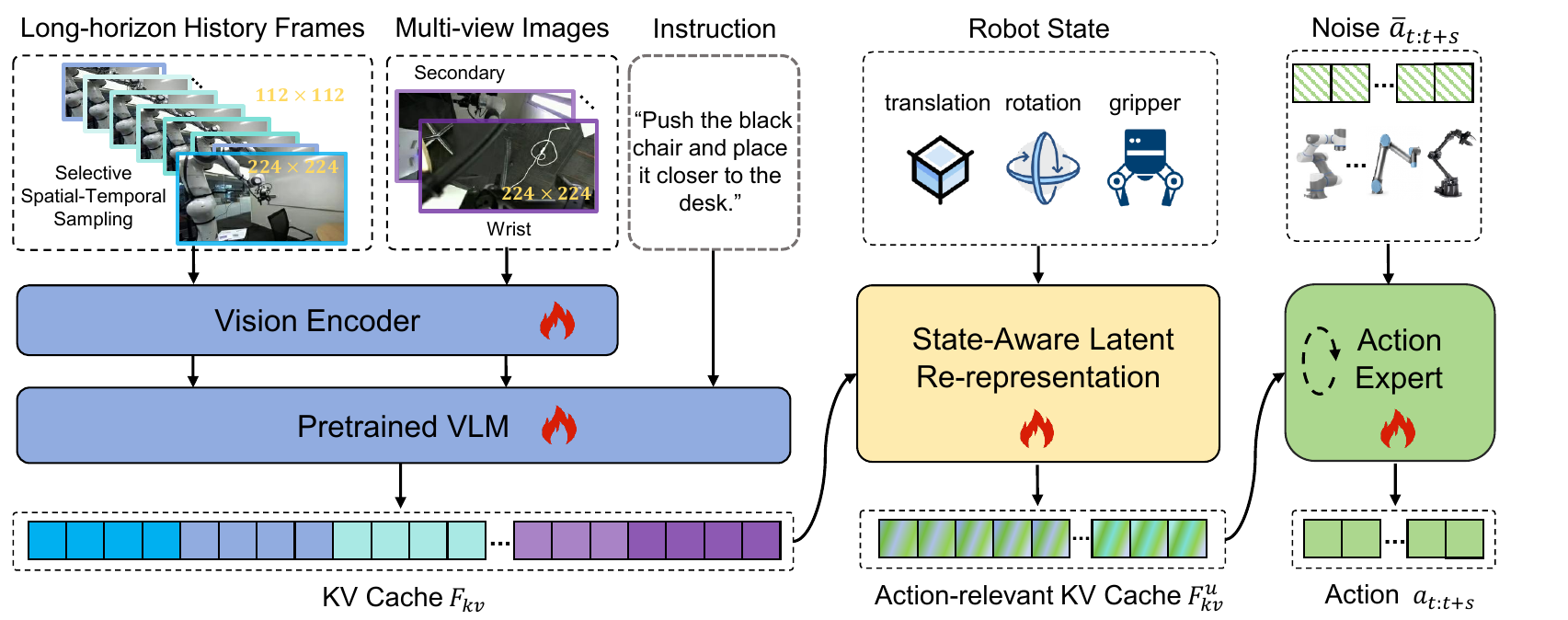}
  \caption{Overview of the proposed model \textbf{LoLA}, which is able to process long-horizon history frames. To align vision-language embeddings with robot actions, the State-Aware Latent Re-representation (SALR) explicitly grounds vision-language features in robot proprioception, creating an ``embodiment-anchored'' latent space conditioned for action generation.
}
  \label{fig:overview}
  
  \hfill
\end{figure*}

\section{Related Work}
\label{sec:formatting}

\subsection{Vision-Language-Action Model}

Recently, relying on the powerful understanding and reasoning capabilities of Vision-Language-Models (VLM), Vision-Language-Action (VLA) models have made rapid progress, which integrates the action generation for adapting the robot manipulation tasks. For example, RT-2 \cite{zitkovich2023rt} fine-tunes the VLM on large-scale vision-language data and robotic demonstration data using next-token prediction. It discretizes robotic actions into 256 binary values and represents them as independent tokens similar to text tokens. OpenVLA \cite{kim2024openvla} adopts a similar discretization approach to fine-tune the Prismatic VLM \cite{karamcheti2024prismatic} on the Open X-Embodiment dataset \cite{vuong2023open}. $\pi_0$ \cite{black2410pi0} consists of a PaliGemma model \cite{beyer2024paligemma} and a separate action expert module, where the VLM is responsible for scene understanding, and the action expert module generates continuous actions through flow matching. Notably, while these models have shown some zero-shot ability, they usually use a single frame and ignore the temporal relationships hindering consecutive action generation and often leading to task failure.

\subsection{Diffusion-based Robot Policy}

Diffusion models \cite{ho2020denoising, song2020denoising, peebles2023scalable}, dominant in image generation, have proven effective in simulating feasible robotic trajectories \cite{chi2023diffusion, reuss2024efficient}. Diffusion policy \cite{chi2023diffusion} represents the visuomotor policy of robots as a conditional denoising diffusion process. Inspired by diffusion policies, Octo \cite{team2024octo} incorporates a small diffusion head with a 3M parameters into a transformer-based backbone architecture to adapt the action outputs of different robots. RDT \cite{liu2024rdt} proposes a pioneering diffusion foundation model for bimanual manipulation, with the diffusion model reaching 1 billion parameters. CogACT \cite{li2024cogact} first uses VLM to generate cognition tokens, then uses them as conditions to guide the diffusion model in generating actions that the robot can understand. However, these methods use vision-language embeddings that are not aligned with actions as conditions to guide action generation. In contrast, our model first aligns the vision-language embeddings with the state information of robots and achieves superior results.

\subsection{Long-Horizon Robot Manipulation}

In the field of robotic manipulation, learning long-horizon tasks has long been a persistent challenge \cite{zhang2024navid, bu2025univla, james2022q, shridhar2023perceiver, feng2025reflective, fan2025diffusion}. These tasks typically involve a series of fine-grained actions, each of which must account for physical constraints and their potential consequences, making them highly challenging for the policy model. For example, a long-horizon task may involve opening a microwave, placing a bowl of milk inside, closing the door, and setting the timer for 10 seconds. When task demonstrations are available, many studies, including PerAct \cite{shridhar2023perceiver}, ARM \cite{james2022q}, and VAPO\cite{borja2022affordance}, attempt to decompose complex long-horizon tasks into multiple stages by identifying sub-goals, thereby providing intermediate learning signals and mitigating the accumulation of action errors. However, these decomposition strategies often rely on task-specific knowledge, making them difficult to generalize to new tasks. Besides, ReflectVLM \cite{feng2025reflective} aims to predict future world states and use these predictions to guide action selection and error correction, while DTP \cite{fan2025diffusion} attempts to adapt to long-horizon tasks by forecasting the trajectories of robots. 
Unlike these methods, our model leverages rich historical information to address long-horizon tasks. Historical frames is more informative, as it includes not only the actions of robots but also the effects of those actions on the environments, such as occlusion relationships caused by the manipulation of robots.

\section{Approach}

Our goal is to develop a generalizable VLA model that enables different robots to accurately perform various long-horizon tasks based on historical frame information, multi-view images, and language instructions. Specifically, given a long-horizon multi-frame input, multi-view images at the current timestep, and a language instruction, the proposed model predicts a temporal action sequence $\{a_{t}, a_{t+1}, a_{t+2}, ..., a_{t+s}\}$ to drive the robot to complete corresponding tasks, where $s$ is the number of predicted future steps. As shown in Figure \ref{fig:overview}, the proposed model consists of three components: the pretrained VLM, the State-Aware Latent Re-representation, and the Action Expert. The pretrained VLM first selects visual embeddings most relevant to the language instruction from long frame sequences and multi-view observations. The State-Aware Latent Re-representation further aligns the vision-language representation with actions, and the Action Expert finally decodes the desired action for the robot from noise based on the aligned embeddings.

\subsection{Pretrained VLM for Visual Encoding}
In this section, we present our visual encoding methodology, which is designed to adapt a pretrained VLM to understand both the instantaneous spatial layout and long-horizon dynamic changes. We achieve this through two specialized encoding components: Current Observation Encoding and Historical Motion Encoding.

\textbf{Current Observation Encoding}: To build a comprehensive and precise spatial understanding of the current scene, this component processes high-fidelity images from the current timestep $t$.
Accurate, high-resolution information of the current state is paramount for precise, low-level action generation process. These inputs include the high-resolution primary view frame $V_t$ and multi-view images $\mathcal{V}_m = \{V_{sec}, ..., V_{wrist} \}$. Each view serves a distinct purpose. The primary view provides the most critical and comprehensive perspective of the task area, while the secondary view from alternative views is crucial for resolving spatial ambiguities and supplementing occluded regions. And the wrist view provides a vital close-up, egocentric perspective of the end-effector and object, which is essential for fine-grained manipulation and alignment. These high-fidelity observations are fed into the vision encoder to extract current observation embeddings $F_{coe}$. This $F_{coe}$ represents a detailed, static snapshot of the current spatial relationship between the robot, objects, and the environment.

\textbf{Historical Motion Encoding:} To capture the dynamic context of the task, understand object motion, and track task progress, this component focuses on processing the long-horizon historical motion embeddings:
\begin{equation}
    \mathcal{V}_{hme}=\{V^{\downarrow}_{t-n}, ..., V^{\downarrow}_{t-3}, V^{\downarrow}_{t-2}, V^{\downarrow}_{t-1} , V_{t}\}.
\end{equation}
We employ a Selective Spatial-Temporal Sampling strategy: all historical frames in $\mathcal{V}_{hme}$ are downsampled to a lower resolution (e.g., $224\times224\xrightarrow{}112\times112$). This design is based on the insight that the immediate action most critically depends on the high-resolution current state, while the history primarily provides context (like velocity, trajectory, or task phase), which can be effectively gleaned from lower-resolution frame sequence. This strategy significantly reduces the computational and memory burden of processing long video sequences while retaining the complementary temporal information.

Accordingly, this downsampled sequence $\mathcal{V}_{hme}$ is fed into the same vision encoder, which processes the frames temporally to extract the historical embeddings $F_{hme}$. Critically, $F_{hme}$ encodes key motion information, but a significant ``modality gap'' still exists between this high-level visual representation and the low-level action generation process. Visual motion (i.e., patterns of changing pixel values) is fundamentally different from physical robot motion (i.e., joint angle velocities, end-effector poses and gripper states). These embeddings are inherently unaligned with the robot's physical proprioception.

To bridge this gap, the embeddings from the current observation $F_{coe}$ and the historical motion $F_{hme}$ are concatenated, along with the language instruction embeddings $F_{l}$, to form the complete input tensor $F_{in}$:

\begin{equation}
    F_{in}= \ \{F_{coe}, F_{hme}, F_{l}\},
\end{equation}
in which the $L$-layer VLM processes this $F_{in}$. At each layer $i$ (from $1$ to $L$), the VLM produces intermediate Key and Value embeddings, $\{K_{i}, V_{i}\}$. These layer-wise VL embeddings $\{K_{i}, V_{i}\}_{i=1}^L$ are the crucial inputs required by our deep fusion mechanism, the State-Aware Latent Re-representation (SALR) module, as described in Section 3.2.

\begin{figure}[t]
  \centering
  \includegraphics[width=1.02\linewidth]{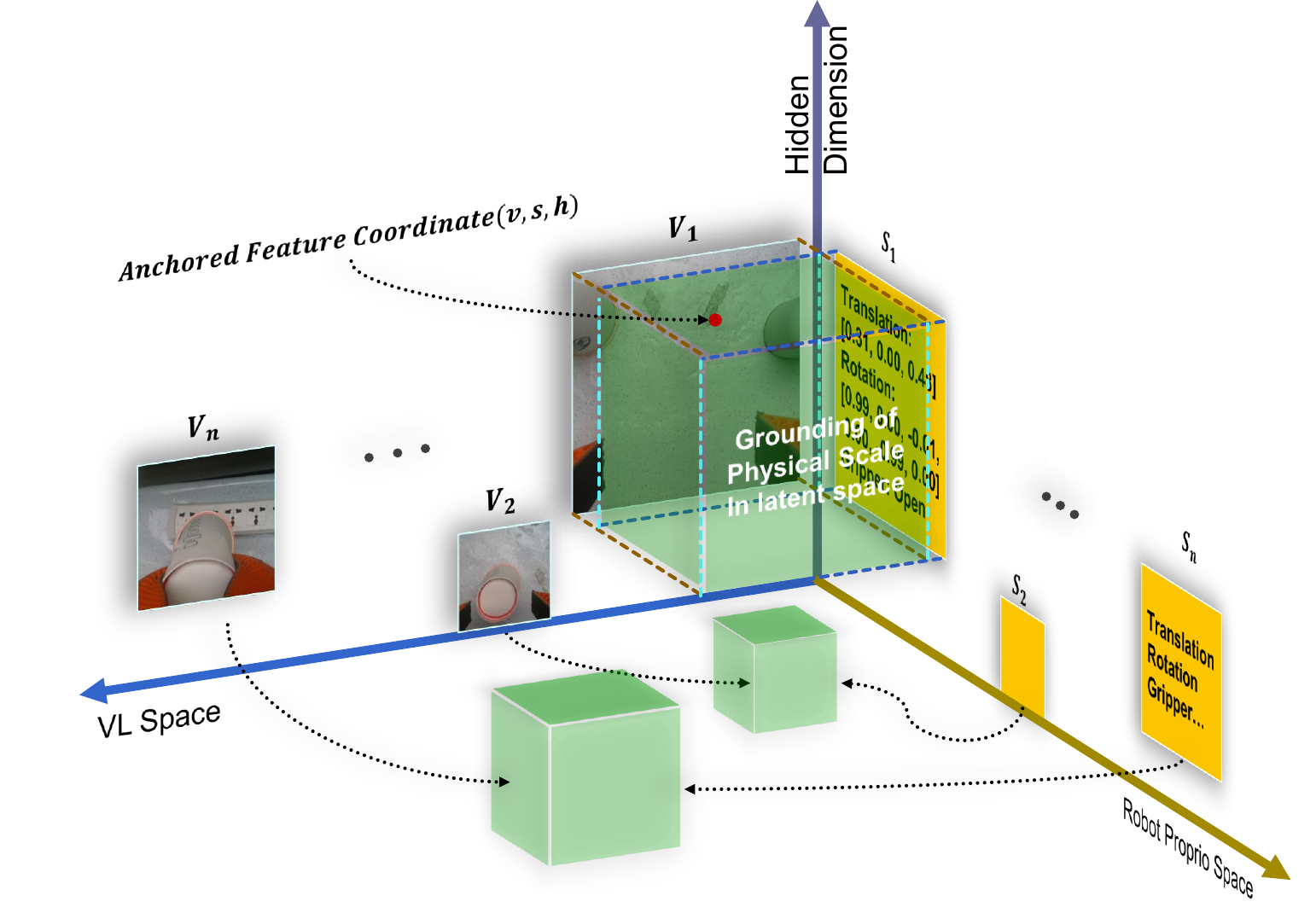}
  \caption{Illustration of the State-Aware Latent Re-representation. The latent space $\textbf{(V, S, H)}$ is formed by outer products between the state embeddings and the key-value from the vision-language embeddings, with a hidden dimension of $h$. The dotted arrows denote projection pairs that map visual observations of robot grippers to real-world physical scales (e.g., translation \& rotation values).
  }
  \label{fig:attn}
\end{figure}

\subsection{State-Aware Latent Re-representation}
In vision-language-action (VLA) models, a significant gap exists between the direct output of vision-language embeddings and actions. 
This discrepancy makes it challenging for the VLA models to directly translate abstract instructions and visual cues into precise robotic actions. Therefore, effectively aligning these modalities is critical to ensure that the robot can correctly understand the task and perform accurate, goal-directed behaviors. However, current methods \cite{black2410pi0, li2024cogact, wen2025tinyvla} suffer from weak alignment between actions and vision-language embeddings. While they often concatenate robot proprioception (e.g., joint angles) with the vision-language embeddings, this simple ``late fusion'' or ``additive'' approach treats the state as just another parallel conditional input.  The vision-language embeddings are not fundamentally transformed or filtered by the physical context. These embeddings remain abstract and ungrounded. This forces the subsequent attention layers to implicitly learn this complex physical grounding from a simple concatenation, which may lead to irrelevant information in the embeddings (e.g., background features) misguiding the action generation. \par

In stark contrast to these abstract embeddings, the robot state $s_t$ (e.g., containing joint angles $q_t$ and end-effector pose $p_t$) and the action $a_t$ are intrinsically coupled and operate in the same physical domain. Unlike abstract visual features, the action is often explicitly defined relative to the state. For instance, in many common control paradigms such as delta-pose or velocity control, the action $a_t$ dictates the change that produces the next state $s_{t+1}$, a relationship that can be expressed as:
\begin{equation}
    s_{t+1} = s_{t} \oplus a_{t},
\end{equation}
where $\oplus$ denotes a composition operation (e.g., pose transformation) in the state space. This direct, physically-grounded relationship makes $s_t$ the ideal ``grounding context'' to filter the ungrounded VL embeddings. 

Therefore,  we propose the \textbf{State-Aware Latent Re-representation (SALR)}, a novel deep fusion mechanism. Instead of a simple late-fusion, we introduce a dedicated \textit{State Transformer} that runs in parallel to the VLM and shares the same $L$-layer depth. The SALR operation occurs at every layer of this State Transformer, replacing standard cross-attention with a state-grounded multiplicative fusion.

Specifically, the process begins by projecting the raw robot state vector $s_t$ through a state projection layer to match the hidden dimension $H$, creating the initial state embedding. This initial state embedding is then fed into the State Transformer for further processing.

Inside each layer $i$, the input state embedding first passes through a self-attention mechanism to produce a refined state representation. Then this representation is projected to generate the robot state's Query vector, $Q_r \in \mathbb{R}^{N_s \times H}$, where $N_s$ denotes the number of attention heads of State Transformer. Instead of a standard dot-product attention, $Q_r$ is used to modulate the VLM's KV cache of each embedding \{$K_i, V_i  \in \mathbb{R}^{N_v \times H}$\} from the corresponding $i$-th layer, using the outer-product fusion. As shown in Figure. \ref{fig:attn}, this computes the fused representations:
\begin{equation}
    K^*[i,j,:]=Q_r[i,:] \odot K_i[j,:],
\end{equation}
\begin{equation}
    V^*[i,j,:]=Q_r[i,:] \odot V_i[j,:].
\end{equation}
This expands the feature space to $(N_s \times N_v \times H)$, where $\odot$ denotes the element-wise product, $N_v$ denotes the number of key value heads in VLM.

To extract action-relevant cues from the latent space, we further introduce a learnable mask for representation in the latent space, which adaptively determines how much information to retain. Formally, this process can be written as: 
\begin{equation}
    \begin{aligned}
        K' &= K^* \odot M_k, \\
        V' &= V^* \odot M_v, \\
    \end{aligned}
\end{equation}
where $M_k \in \mathbb{R}^{(N_s \times N_v) \times H}$ and $M_v \in \mathbb{R}^{(N_s \times N_v) \times H}$ denotes the learnable mask for key and value's representation in the latent space, respectively. Finally, to further compress the representation space, we utilize a latent space compression strategy to obtain re-encoded key embeddings $K^a \in \mathbb{R}^{N_a \times H}$ and value embeddings $V^a \in \mathbb{R}^{N_a \times H}$, where $N_a$ denotes the number of key value heads of action expert.
\subsection{Action Expert for Action Prediction}
The Action Expert is responsible for translating the high-level, physically-grounded representations from the SALR module into a concrete, multi-step action sequence. Following recent successes in trajectory modeling \cite{black2410pi0}, we implement the expert as a Conditional Flow Matching (CFM) model. The underlying network architecture is an $M$-layer Transformer Decoder. This decoder is designed to be conditioned on three distinct inputs: 1) the final aligned latent condition \{$K^a, V^a$\} from the SALR module, 2) a noisy action trajectory  sampled at time $z$, and 3) The noise timestep $z$, which is encoded via sinusoidal embeddings.

This architecture explicitly separates modalities. The noisy actionand $z$ embeddings are processed by the decoder's Self-Attention layers to model the temporal structure of the action sequence. At each of its $M$ layers, a Cross-Attention mechanism injects the state-aware, multi-modal guidance by attending to the $K^a$ (as Key) and $V^a$ (as Value). During training, the expert is optimized using the CFM objective to predict the noise. During inference, it takes the $\{K^a, V^a\}$ condition and performs multiple denoising steps to progressively decode a smooth and precise action sequence from pure noise.
\begin{figure*}[h]
  \centering
  
  \includegraphics[width=0.85\textwidth]{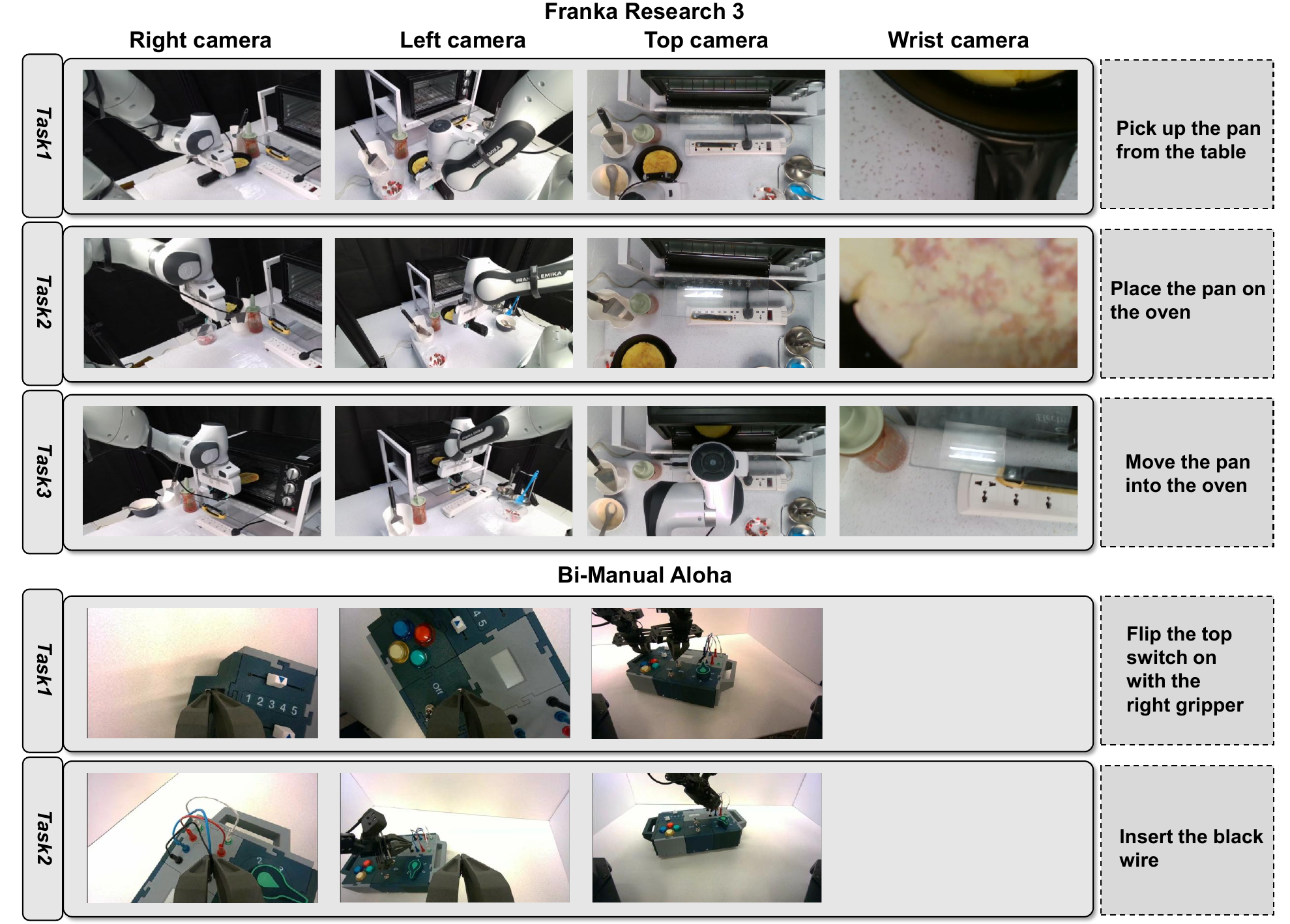}
  \caption{Real-world experimental setup featuring Franka Research 3 and bi-manual Aloha robots, displaying multi-view camera inputs and corresponding text instructions. Specifically, Tasks 1, 2, and 3 form a sequential, long-horizon manipulation task: \textit{"Put the flat-bottomed pan from the table onto the oven platform"}.}
  \label{fig:real_world}
\end{figure*}

\section{Experiments}
\label{sec:exp}
\begin{table*}[h]
    \centering
    \caption{Comparison of our approach with existing VLA models across four tasks in two SIMPLER settings on the Google robot.}
    \label{tab:simpler_google}
    
    \begin{tabular}{c|c|cccc|c}
        \toprule

        \textbf{\shortstack{Google \\ Robot}} & \textbf{Method} & \shortstack{Pick \\ Coke Can} & \shortstack{Move \\ Near} & \shortstack{Open/Close \\ Drawer} & \shortstack{Open Top Drawer \\ and Place Apple} & Average \\
        
        
        \midrule
        \multirow{6}{*}{\shortstack{Visual \\ Matching}} 
        & RT-1 \cite{brohan2022rt} & 85.7\% & 44.2\% & \textbf{73.0\%} & 6.5\% & 52.4\% \\
        & RT-1-X \cite{vuong2023open} & 56.7\% & 31.7\% & 59.7\% & 21.3\% & 42.4\% \\
        & RT-2-X \cite{vuong2023open} & 78.7\% & \textbf{77.9\%} & 25.0\% & 3.7\% & 46.3\% \\
        & Octo-Base \cite{team2024octo} & 17.0\% & 4.2\% & 22.7\% & 0.0\% & 11.0\% \\
        & OpenVLA \cite{kim2024openvla} & 18.0\% & 56.3\% & 63.0\% & 0.0\% & 34.3\% \\
        & $\pi_0$ \cite{black2410pi0} & 87.3\% & 35.0\% & 72.6\% & 16.0\% & 52.7\% \\
        \midrule
        & LoLA (Ours) & \textbf{88.0\%} & 71.7\% & 59.3\% & \textbf{26.9}\% & \textbf{61.5\%} \\
        \midrule
        \multirow{6}{*}{\shortstack{Variant \\ Aggregation}} 
        & RT-1 \cite{brohan2022rt} & \textbf{89.8\%} & 50.0\% & 32.3\% & 2.6\% & 43.7\% \\
        & RT-1-X \cite{vuong2023open} & 49.0\% & 32.3\% & 29.4\% & 10.1\% & 30.2\% \\
        & RT-2-X \cite{vuong2023open} & 82.3\% & \textbf{79.2\%} & 35.3\% & \textbf{20.6\%} & 54.4\% \\
        & Octo-Base \cite{team2024octo} & 0.6\% & 3.1\% & 1.1\% & 0.0\% & 1.2\% \\
        & OpenVLA \cite{kim2024openvla} & 60.8\% & 67.7\% & 28.3\% & 1.2\% & 39.3\% \\
        & $\pi_0$ \cite{black2410pi0} & 85.2\% & 40.8\% & 42.1\% & 15.9\% & 46.0\% \\
        \midrule
        & LoLA (Ours) & 83.2\% & 56.7\% & \textbf{63.2\%} & 15.3\% & \textbf{54.6\%} \\
        \bottomrule
    \end{tabular}
    
\end{table*}

\subsection{Implementation Details}
Our LoLA model is pretrained on the OXE dataset \cite{o2024open} and the AgiBot dataset \cite{bu2025agibot}, comprising 1.1 million real-world robot episodes (detailed sampling rates are in the supplementary materials). Our architecture can utilize VLMs with multi-frame or video processing capabilities, such as Gemma3 \cite{gemma_2025} or Qwen2.5-VL \cite{bai2025qwen2}. We use the latter for its robust video understanding performance. The model is trained end-to-end on a cluster of 32 NVIDIA A100 GPUs with batch size of 1280. By default, 25 historical frames are used to make a balance between information capacity and compute efficiency. We evaluate LoLA across simulation benchmarks (SIMPLER \cite{li2024evaluating}, LIBERO \cite{liu2023libero}) and real-world scenarios (Franka and Bi-Manual Aloha platforms \cite{busybox2025}). 

\subsection{Manipulation Benchmark on SIMPLER}
We evaluate LoLA on the SIMPLER \cite{li2024evaluating} benchmark, which is designed to test sim-to-real generalization by replicating real-world scenarios for the Google Robot and WidowX Robot under ``Visual Matching'' and ``Variant Aggregation'' settings. The ``Visual Matching'' setting aims to reduce the visual appearance gap between real environments and raw simulation by overlaying real-world images onto simulation backgrounds, while the ``Variant Aggregation'' setting creates different simulation environment variants based on Visual Matching.

As shown in Table \ref{tab:simpler_google}, on the Google Robot tasks, LoLA achieves state-of-the-art performance. It significantly outperforms the $\pi_0$ baseline across both settings and surpasses the performance of the larger, closed-source RT-2-X model.

As shown in Table \ref{tab:simpler_widowx}, the advantages of our approach are even more pronounced on the complex, multi-step tasks of the WidowX robot. LoLA achieves a 71.9\% average success rate, demonstrating a substantial 20.6\% relative improvement over $\pi_0$. This highlights our model's effectiveness in leveraging historical context to maintain coherence in long-horizon tasks.
\begin{table*}[h]
    \centering
    \caption{Comparison with existing VLA models across four tasks in the SIMPLER (Visual Matching) setting on the WidowX robot.}
    \label{tab:simpler_widowx}
    
    \begin{tabular}{c|c|cccc|c}
        \toprule
        
        \textbf{\shortstack{WidowX \\ Robot}} & \textbf{Method} & \shortstack{Put Spoon \\ on Towel} & \shortstack{Put Carrot \\ on Plate} & \shortstack{Stack Green Block \\ on Yellow Block} & \shortstack{Put Eggplant \\ in Yellow Basket} & Average \\
        
        \midrule
        \multirow{7}{*}{\shortstack{Visual \\ Matching} }
        & RT-1-X \cite{brohan2022rt} & 0.0\% & 4.2\% & 0.0\% & 0.0\% & 1.1\% \\
        & Octo-Base \cite{team2024octo}  & 15.8\% & 12.5\% & 0.0\% & 41.7\% & 17.5\% \\
        & Octo-Small \cite{team2024octo} & 41.7\% & 8.2\% & 0.0\% & 56.7\% & 26.7\% \\
        & OpenVLA \cite{kim2024openvla} & 4.2\% & 0.0\% & 0.0\% & 12.5\% & 4.2\% \\
        & $\pi_0$ \cite{black2410pi0} & 62.5\% & \textbf{66.7\%} & 25.0\% & 12.5\% & 41.7\% \\
        & SpatialVLA \cite{qu2025spatialvla} & 16.7\% & 25.0\% & 29.2\% & \textbf{100\%} & 42.7\% \\
        & CogACT \cite{li2024cogact} & 71.7\% & 50.8\% & 15.0\% & 67.5\% & 51.3\%\\
        \midrule
        & LoLA (Ours) & \textbf{95.8$\%$} & 58.3$\%$ & \textbf{54.2$\%$} & 79.2$\%$ & \textbf{71.9$\%$} \\
        \bottomrule
    \end{tabular}
    
\end{table*}
\begin{table*}[h]
    \centering
    \caption{Comparison of our approach with existing VLA models on the four LIBERO simulation environments.}
    \label{tab:libero_results}
    \begin{tabular}{c|cccc|c}
        \toprule
        \textbf{Method} & LIBERO-Goal & LIBERO-Object & LIBERO-Spatial & LIBERO-Long & Average \\
        \midrule
        Diffusion Policy \cite{chi2023diffusion} & 68.3\% & 92.5\% & 78.3\% & 50.5\% & 72.4\% \\
        Octo \cite{team2024octo} & 84.6\% & 85.7\% & 78.9\% & 51.1\% & 75.1\% \\
        OpenVLA \cite{kim2024openvla} & 79.2\% & 88.4\% & 84.7\% & 53.7\% & 76.5\% \\
        TraceVLA \cite{zheng2024tracevla} & 75.1\% & 85.2\% & 84.6\% & 54.1\% & 74.8\% \\
        RDT \cite{liu2024rdt} & 68.2\% & 77.8\% & 60.2\% & 29.0\% & 58.8\% \\
        $\pi_0$ \cite{black2410pi0} & 94.0\% & 97.8\% & 91.4\% & 85.4\% & 92.2\% \\
        \midrule
        LoLA (Ours) & \textbf{97.2\%} & \textbf{99.6\%} & \textbf{99.6\%} & \textbf{88.2\%} & \textbf{96.2\%} \\

        \bottomrule
    \end{tabular}

\end{table*}
\begin{table*}[h]
    \centering
    \caption{Comparison of our approach with existing VLA models on Single-Step tasks in real-world scenarios.}
    \label{tab:real_results_single_step}
    \small
    \begin{tabular}{c|cccccccccc}
        \toprule
        \textbf{Method} & T1 & T2 & T3 & T4 & T5 & T6 & T7 & T8 & T9 & Avg\\
        \midrule
         DP \cite{chi2023diffusion} & 23.1\% & 0.0\% & 23.1\%  & 30.8\%  & 23.1\%  & 7.7\% & 0.0\% & 30.8\% & 23.1\% & 18.0\%\\
         $\pi_0$ \cite{black2410pi0} & \textbf{46.2\%}  & 38.5\% & 38.5\%  & 53.8\%  & 23.1\%  & 23.1\% & 30.8\%  & 53.8\% & 46.2\% & 36.8\%\\
         \midrule
         
         LoLA (Ours) & 15.4\% & 38.5\%  & \textbf{53.8\%}  & \textbf{61.5\%}  & \textbf{61.5\%}  & 15.4\% & \textbf{53.8\%} & 53.8\% & \textbf{61.5\%} & \textbf{46.1\%} \\
        \bottomrule
    \end{tabular}
\end{table*}

\begin{table*}[t]
    \centering
    \caption{Impact of each component. \textbf{FrozenVL}, \textbf{MF}, and \textbf{SALR} denote freezing the pretrained VLM weights, multiple historical frames, and the State-Aware Latent Re-representation, respectively.}
    \label{tab:ablation}
    \begin{tabular}{ccc|cccc|c}
    \toprule
          FrozenVL  & MF & SALR & \shortstack{Put Spoon \\ on Towel}  & \shortstack{Put Carrot \\ on Plate} & \shortstack{Stack Green Block \\ on Yellow Block} & \shortstack{Put Eggplant \\ in Yellow Basket} & Average \\
    \midrule
         & & & 41.7\% & 50\% & 12.5\% & 16.7\% & 30.3\% \\
        & \checkmark & & 54.2\% & 45.8\% & 37.5\% & 29.2\% & 41.7\% \\
         & & \checkmark & 66.7\% & 50.0\% & 16.7\% & 45.8\% & 44.8\% \\
        & \checkmark & \checkmark & 54.2\% & 41.7\% & 54.2\% & 79.2\% & 57.3\% \\
         \checkmark &  \checkmark & \checkmark & 4.2\% & 0 & 0 & 0 & 1.1\% \\
    \bottomrule
    \end{tabular}
\end{table*}

\subsection{Manipulation Benchmark on LIBERO}
We evaluate our model on the LIBERO \cite{liu2023libero} benchmark, which assesses generalization across diverse objects, goals, spatial relationships and long-horizon tasks. The LIBERO benchmark consists of four task suites, which are \textit{LIBERO-Goal}, \textit{LIBERO-Spatial}, \textit{LIBERO-Object} and \textit{LIBERO-Long}, each comprising 10 tasks with 50 human-teleoperated demonstrations. Specifically, the tasks in \textit{LIBERO-Spatial}, \textit{LIBERO-Object}, and \textit{LIBERO-Long} are generally formulated as \textbf{Multi-Step} tasks. Their instructions often follow an “A and B” format, inherently decomposing into multiple sub-tasks. For example, in \textit{LIBERO-Object}, the task pick up the milk and place it in the basket requires grasping the target object first and then placing it at a designated location, thus involving sequential actions.

As shown in table \ref{tab:libero_results}, LoLA achieves the highest average success rate (96.2\%) across all task suites, surpassing existing state-of-the-art methods, including $\pi_0$. Notably, our model demonstrates robust performance on the challenging LIBERO-Long suite (88.2\%), which specifically tests long-horizon capabilities. This result underscores our model's ability to effectively leverage historical information to guide the robot in complex, multi-step tasks.
\begin{table}[b]
    \centering
    \caption{Comparison of our approach with existing VLA models on Multi-Step tasks in real-world scenarios using the Franka Robot. The percentages indicate the success rate for models completing at least two subtasks.}
    \label{tab:real_results_multi_step}
    \small
    \begin{tabular}{c|ccc}
        \toprule
        \textbf{Method} & T1 , T2 , T3 & T4 , T5 , T6 & T7 , T8 , T9 \\
        \midrule
         DP \cite{chi2023diffusion} & 0.0\% & 7.1\% & 0.0\%  \\
         $\pi_0$ \cite{black2410pi0} & \textbf{17.8\%}  & 12.4\% & 16.6\%  \\
         \midrule
         LoLA (Ours) & 5.9\% & \textbf{33.1\%}  & \textbf{28.9\%}   \\
        \bottomrule
    \end{tabular}
    
\end{table}

\subsection{Real-World Evaluation}
To validate LoLA in non-simulated environments, we conduct real-world evaluations on two platforms: a Franka Research 3 robot setup and Bi-Manual Aloha setup \cite{busybox2025}.

Our real-world evaluation utilizes two distinct setups. First, for the Franka platform, we collected a benchmark comprising 28 tasks using a real-time teleoperation system (comprising a 7-DoF arm and 1-DoF gripper) that operates at up to 50Hz. Specifically, we designed 22 fine-grained atomic tasks cover a variety of common manipulation actions involved in pizza preparation, such as spreading sauce, cutting pizza, and placing the pan into a microwave oven for cooking. These atomic tasks are logically grouped into 7 sequential long-horizon tasks (each consisting of 2-5 sub-tasks) to evaluate multi-stage state transitions. Furthermore, to test the limit of temporal consistency without intermediate resets, we collected 6 end-to-end long-horizon tasks, where the robot performs long-horizon work continuously. In total, we have collected 13 long-horizon tasks with an average duration of approximately 2.3 minutes. Second, we use the BusyBox Bi-Manual Aloha setup \cite{busybox2025}, which features six 3D-printable, reconfigurable modules (buttons, switches, etc.). This setup tests manipulation based on physical features, not memorized positions, using the 61 canonical tasks from the Busybox dataset. The Busybox dataset consists of 1253 human-teleoperated demonstrations of which 18.7\% is longer than 20 seconds.

For the experiment, we choose 9 tasks for real-world evaluation: \textit{Pick up the pan from the table} (\textbf{T1}), \textit{Place the flat-bottomed pan on the oven} (\textbf{T2}), \textit{Move the pan into the oven} (\textbf{T3}), \textit{Push the flat-bottomed pan into the oven} (\textbf{T4}), \textit{Close the oven} (\textbf{T5}), \textit{Rotate the oven dial} (\textbf{T6}),  \textit{Open the oven} (\textbf{T7}),  \textit{Remove the flat-bottomed pan from the oven} (\textbf{T8}),  \textit{Return the pan from the oven to the table} (\textbf{T9}). We further design three long-horzion tasks: \textit{Put the flat-bottomed pan from the table onto the oven platform} ($\textbf{T1} \rightarrow \textbf{T2} \rightarrow \textbf{T3}$), \textit{Set up the oven} ($\textbf{T4} \rightarrow \textbf{T5} \rightarrow \textbf{T6}$) and \textit{Remove the pan from the oven and place it back on the table} ($\textbf{T7} \rightarrow \textbf{T8} \rightarrow \textbf{T9}$). Visual illustrations for these tasks can be found in Figure \ref{fig:real_world}.

As shown in Table \ref{tab:real_results_single_step}, LoLA outperforms baselines (Diffusion Policy, $\pi_0$) on the average success rate of single-step tasks. The primary challenge, however, lies in the long-horizon sequences. In Table \ref{tab:real_results_multi_step}, we report the success rate for models completing at least two subtasks. In challenging long-horizon scenarios, LoLA demonstrates significant improvements, achieving up to 2.67$\times$ the success rate of $\pi_0$. It is critical to note that this Franka setup is inherently challenging. As shown in Table \ref{tab:real_results_single_step}, the $\pi_0$ baseline achieves less than 40\% average success on the single-step tasks. Given that errors from these difficult individual steps compound, the absolute success rates for all models on the sequential long-horizon tasks are expectedly low. Therefore, the significant relative improvement that LoLA demonstrates in this high-difficulty, error-accumulating real-world setting further validates the robustness of our long-term context integration module SALR. These combined results validate the effectiveness of our SALR module in complex, real-world scenarios. Detailed Busybox Bi-Manual Aloha results are summarized in Supplementary Material.

\subsection{Ablation Study}
\textbf{Components Analysis.}
We conduct ablation experiments on the WidowX robot in the SIMPLER simulated environment and report the average manipulation accuracy. MF denotes the use of multiple historical frames. FrozenVL indicates that the parameters of the VLM remain unchanged. SALR refers to the state-aware latent re-representation. Note that without MF, only the current frame is used; without SALR, a simple MLP is employed to adjust the number of heads. As shown in Table \ref{tab:ablation}, incorporating MF significantly improves the overall manipulation success rate. When both SALR and MF are applied, the model achieves its best performance. These results underscore the importance of leveraging historical frame information and effectively aligning vision-language embeddings with actions.

  
  


\textbf{Impact of the robot proprioception.} We propose to use the robot proprioception vector to extract action-relevant features from vision-language representations. To study the impact of the robot proprioception, we conduct experiments on the LIBERO simulation environment. As shown in Table \ref{tab:libero_results_ab_state}, explicitly grounding the VL embeddings with the robot proprioception ("w/ state") significantly improves the average success rate from 84.7\% to 91.2\%. This 6.5\% gain confirms the critical role of our state-aware alignment module (SALR) for robust performance.
\begin{table}[b]
    \centering \small
    \caption{Performance comparison with and without robot proprioception in the LIBERO simulation environment.}
    \label{tab:libero_results_ab_state}
    
    \begin{tabular}{c|cccc|c}
        \toprule
        \textbf{} & Goal & Object & Spatial & Long & Average  \\
        \midrule
        w/o state & 84.2\% & 90.0\% &  93.0\% & 71.5\% & 84.7\% \\
        w/ \ \ state & \textbf{91.4\%} & \textbf{96.0\%} & \textbf{95.5\%} & \textbf{82.0\%} & \textbf{91.2\%} \\
       
        \bottomrule
    \end{tabular}
\end{table}
\section{Conclusion}
In this paper, we propose LoLA, a Vision-Language-Action model designed to address the challenge of long-horizon robotic manipulation. To balance the high computational cost of long image sequence inputs with the need for rich temporal context, our approach employs a selective spatial-temporal sampling strategy. This method efficiently processes both high-fidelity current observations and downsampled historical motion context.

The core of our contribution is the State-Aware Latent Re-representation (SALR), a novel deep fusion mechanism designed to bridge the critical modality gap. Unlike prior methods using simple concatenation or ``late fusion'', SALR introduces a dedicated State Transformer that runs in parallel to the VLM. At each layer, SALR explicitly grounds the VLM's intermediate embeddings via a state-grounded multiplicative outer-product fusion. These refined representations then condition a Conditional Flow Matching (CFM) Action Expert to generate smooth, multi-step actions.

Extensive experiments on simulation (SIMPLER, LIBERO) and real-world setup (Franka, Bi-manual Aloha) benchmarks demonstrate that LoLA significantly outperforms existing VLA models. The performance gains are particularly pronounced in long-horizon and error-prone sequential tasks, validating our architecture's ability to maintain coherence and physical grounding over time. While our real-world experiments validate LoLA's effectiveness, a limitation remains in achieving robust execution for more complex, novel, or perturbation-rich long-horizon tasks. Future work will focus on enhancing this robustness, for example, by enabling dynamic, closed-loop recovery from significant failures. This will further push the boundaries of robust long-horizon task execution in challenging real-world environments. 

\section*{Acknowledgement}
This work was supported in part by Science and Technology Innovation (STI) 2030---Major Projects under Grant 2022ZD0208700, and National Natural Science Foundation of China under Grant 62376264.

{
    \small
    \bibliographystyle{ieeenat_fullname}
    \bibliography{main}
}
\clearpage
\setcounter{page}{1}
\maketitlesupplementary

\section{Implementation Details}
\label{sec:arch_detail}
In this section, we provide a comprehensive overview of the implementation details, including the composition of the pre-training dataset, the configuration of the real-world Franka environment, model architecture specifications, data collection protocols, and the distributed training infrastructure used to train our model.

\subsection{Training Data Details}
We pretrained LoLA on the OXE~\cite{o2024open} and AgiBot datasets~\cite{bu2025agibot}. Our data mixture strategy primarily follows~\cite{team2024octo, vuong2023open}. The detailed data mixture is listed in Table~\ref{tab:dataset_mix}. We use 1.0 million robot trajectories to pretrain LoLA, containing approximately 62 million timestamps.

\begin{table}[h]
    \caption{Mixture of datasets used during pretraining, including OXE \cite{o2024open} and AgiBoT \cite{bu2025agibot}.}
    \label{tab:dataset_mix}
    \centering
    \begin{tabular}{l r}
        \toprule
        \textbf{Training Dataset Mixture} & \\
        \midrule
        Fractal \cite{brohan2022rt} & 14.3\% \\
        Kuka \cite{kalashnikov2018scalable} & 14.4\% \\
        Bridge \cite{ebert2021bridge, walke2023bridgedata} & 13.3\% \\
        Taco Play \cite{mees2022grounding, rosete2023latent} & 3.0\% \\
        Jaco Play \cite{dass2023jacoplay} & 0.5\% \\
        Berkeley Cable Routing \cite{luo2024multistage} & 0.3\% \\
        Roboturk \cite{mandlekar2019scaling} & 2.4\% \\
        Viola \cite{zhu2023viola} & 1.0\% \\
        Berkeley Autolab UR5 \cite{BerkeleyUR5Website} & 1.2\% \\
        Toto \cite{zhou2023train} & 2.1\% \\
        Language Table \cite{lynch2023interactive} & 5.0\% \\
        Stanford Hydra Dataset \cite{belkhale2023hydra} & 5.0\% \\
        Austin Buds Dataset \cite{zhu2022bottom} & 0.2\% \\
        NYU Franka Play Dataset \cite{cui2022play} & 0.7\% \\
        Furniture Bench Dataset \cite{heo2025furniturebench} & 2.8\% \\
        UCSD Kitchen Dataset \cite{ucsd_kitchens} & $<0.1$\% \\
        Austin Sailor Dataset \cite{nasiriany2022learning} & 2.5\% \\
        Austin Sirius Dataset \cite{liu2025robot} & 2.0\% \\
        DLR EDAN Shared Control \cite{quere2020shared} & $<0.1$\% \\
        IAMLab CMU Pickup Insert \cite{saxena2023multi} & 1.0\% \\
        UTAustin Mutex \cite{shah2023mutex} & 2.5\% \\
        Berkeley Fanuc Manipulation \cite{zhu2023fanuc} & 8.8\% \\
        CMU Stretch \cite{nasiriany2022learning} & 0.2\% \\
        BC-Z \cite{jang2022bc} & 7.7\% \\
        FMB Dataset \cite{luo2025fmb} & 8.9\% \\
        DobbE \cite{shafiullah2023bringing} & 1.6\% \\
        AgiBoT \cite{bu2025agibot} & 7.1\% \\
        \bottomrule
    \end{tabular}
\end{table}

\subsection{Real-World Franka Setup and Data Collection}
\textbf{Hardware Configuration.} Our real-world long-horizon experiments are conducted on a Franka Emika Research 3 robot arm (7-DoF) equipped with a 1-DoF parallel-jaw gripper. The visual perception system consists of four RGB cameras streaming at 30Hz.

\begin{figure*}[h]
    \centering
    \resizebox{\textwidth}{!}{
        \includegraphics[]{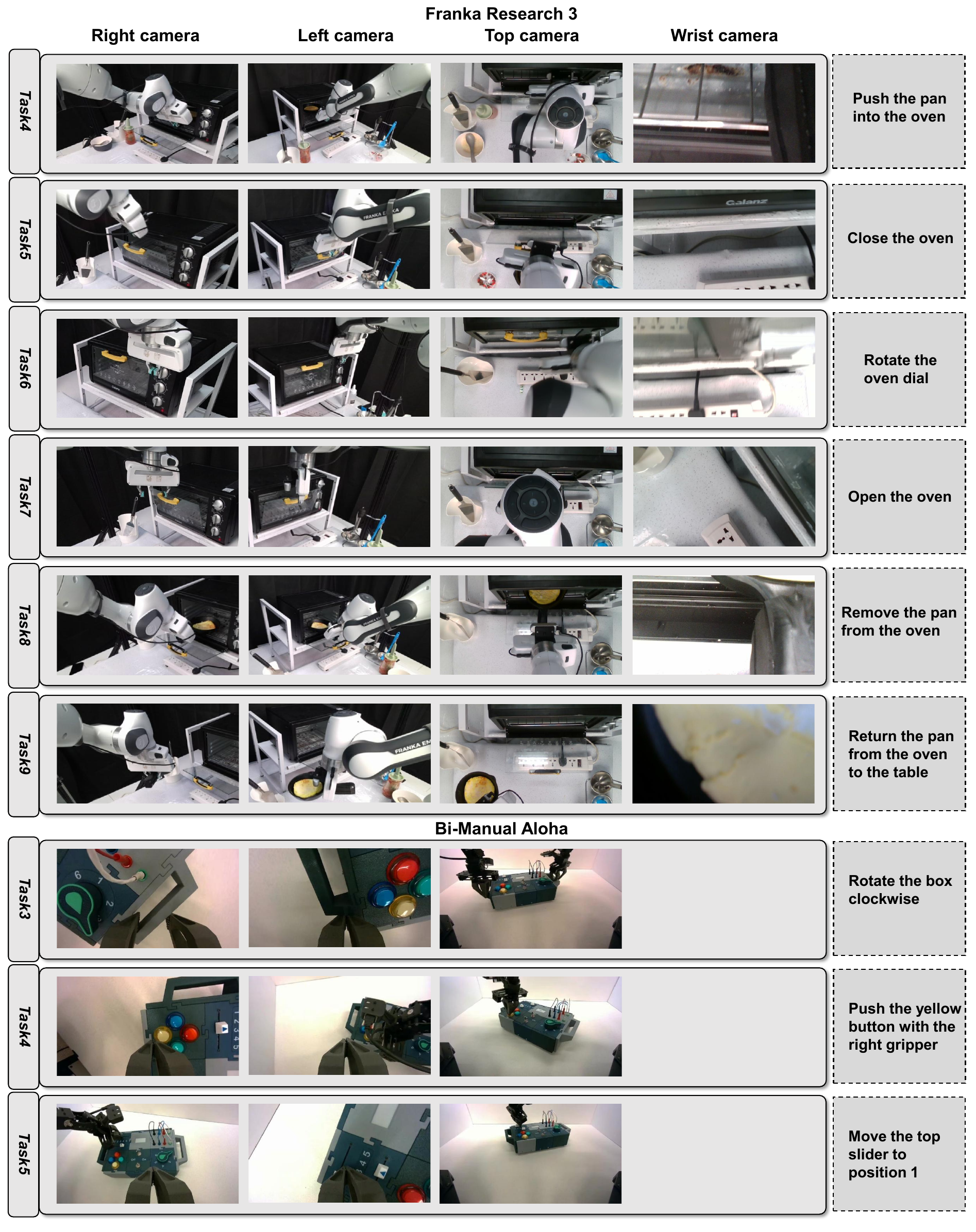}
    }
    
    \caption{\textbf{Visualization of Representative Task Scenarios.} \textbf{Top:} Six fine-grained atomic sub-tasks in the Franka long-horizon cooking task organized into 2 sequential long-horizon tasks, Set up the oven and Remove the pan from the oven and place it back on the table. \textbf{Bottom:} Three distinct tasks from the BusyBox benchmark.}
    \label{fig:additional_tasks}
\end{figure*}

The primary view (right) captures the global workspace from the right side. The secondary view (left) provides a complementary perspective to handle occlusions. Top-Down View is mounted directly above the table for spatial alignment. The wrist view is mounted on the end-effector for fine-grained interaction details.All images are resized to $224 \times 224$ for the current observation and $112 \times 112$ for historical frames before being fed into the model. Figure~\ref{fig:additional_tasks} (Top) visualizes six representative interaction scenarios. \\ \textbf{Data Collection Protocol.}
To collect all of the cooking-themed tasks (e.g., \textit{oven operation}, \textit{pan manipulation}), we developed a low-latency teleoperation system. A human operator controls the Franka robot using a wireless Xbox controller. The control mapping is designed for intuitive 6-DoF manipulation:

\begin{itemize}
    \item Gripper Control: The \textbf{A} button toggles the gripper open and closed.
    \item XY-Plane Translation \& Rotation: The left joystick controls translation in the XY plane, while the right joystick controls rotation around the XY axes (roll and pitch).
    \item Z-Axis Control: The left and right triggers control translation along the Z-axis (up/down), and the left and right bumpers (shoulder buttons) control rotation around the Z-axis (yaw).
\end{itemize}

This setup allows for smooth, continuous control of the 7-DoF arm. The system records proprioceptive states (joint angles, end-effector position and gripper width) and synchronized video streams directly at 20Hz. This frequency was chosen to ensure data quality and stability, preventing frame drops or missing robot states. \\ \textbf{Task Composition.} To evaluate long-horizon manipulation capabilities, we constructed a benchmark comprising 28 tasks. These tasks are structured into two distinct categories to test different aspects of temporal coherence:

\begin{itemize}
    \item Sequential Long-Horizon Groups: We defined 22 fine-grained atomic sub-tasks which serve as building blocks. These are logically organized into 7 sequential long-horizon tasks, each consisting of 2 to 5 consecutive steps, designed to assess the model's ability to handle multi-stage state transitions.
    \item End-to-End Episodes: To test robustness against accumulated errors without intermediate resets, we collected 6 end-to-end continuous tasks (with an average duration of 3.6 minutes), where the robot executes complete cooking workflows continuously. The detailed composition of these long-horizon tasks is listed in Table
\end{itemize}

\begin{table}[h]
    \centering
    \caption{Success rates on the Bi-Manual Aloha BusyBox benchmark. We report the average success rate across 6 of all the canonical tasks. LoLA demonstrates superior generalization capabilities compared to baselines.}
    \label{tab:busybox_results}
    \begin{tabular}{l|c}
        \toprule
        \textbf{Method} & Average Success Rate (\%) \\
        \midrule
        Diffusion Policy \cite{chi2023diffusion} & 8.3\% \\
        $\pi_0$ \cite{black2410pi0} & 30.0\% \\
        \midrule
        \textbf{LoLA (Ours)} & \textbf{46.7\%} \\
        \bottomrule
    \end{tabular}
\end{table}

\begin{table*}[t]
\centering
\caption{Detailed composition of the Franka Benchmark. It includes 7 Sequential Groups (composed of 22 atomic sub-tasks) and 6 End-to-End Episodes.}
\label{tab:franka_tasks}
\begin{tabular}{c|l|c}
\toprule
\textbf{ID} & \textbf{Task Description / Sequence} & \textbf{Duration} \\
\midrule
\multicolumn{3}{c}{\textit{\textbf{Sequential Groups} (22 Atomic Tasks in 7 Groups)}} \\
\midrule
G1 & \textbf{Oven Prep}: Pick pan $\rightarrow$ Place on oven $\rightarrow$ Place into oven & $\sim$1.0 min \\
G2 & \textbf{Oven Ops}: Push pan inward $\rightarrow$ Close oven door $\rightarrow$ Rotate dial & $\sim$1.5 min \\
G3 & \textbf{Serving}: Open oven door $\rightarrow$ Remove pan $\rightarrow$ Place on table & $\sim$1.1 min \\
G4 & \textbf{Cutting}: Grasp cutter $\rightarrow$ Cut pizza $\rightarrow$ Return cutter & $\sim$1.0 min \\
G5 & \textbf{Saucing}: Pick brush $\rightarrow$ Spread sauce $\rightarrow$ Return brush & $\sim$1.0 min \\
G6 & \textbf{Topping}: Pick bowl $\rightarrow$ Pour toppings $\rightarrow$ Return bowl $\rightarrow$ Pick spoon $\rightarrow$ Sprinkle cheese & $\sim$1.7 min \\
G7 & \textbf{Power Supply}: Plug into outlet $\rightarrow$ Switch on & $\sim$1.2 min \\
\midrule
\multicolumn{3}{c}{\textit{\textbf{End-to-End Episodes} (Continuous Execution)}} \\
\midrule
E1 & \textbf{Full Baking Cycle}: Load, bake, and retrieve pizza & $\sim$5.2 min \\
E2 & \textbf{Sauce \& Topping}: Spread sauce and add toppings & $\sim$4.0 min \\
E3 & \textbf{Block Insertion}: Insert multiple blocks into target slots & $\sim$4.8 min \\
E4 & \textbf{Food Preparation}: Pour cheese powder from bottle to bowl & $\sim$2.5 min \\
E5 & \textbf{Hidden Block Retrieval}: Locate and lift the cup hiding the block & $\sim$1.4 min \\
E6 & \textbf{Kitchen Cleanup}: Return tools to storage and power off & $\sim$3.6 min \\
\bottomrule
\end{tabular}%
\end{table*}

\subsection{Model Architecture Specifications}
\label{sec:model_arch}

LoLA is a large-scale VLA model with a total of approximately \textbf{10 Billion parameters}. The architecture consists of three main components, whose specifications are detailed below:

\begin{itemize}
    \item Vision-Language Backbone: We initialize our backbone with Qwen2.5-VL-7B, which serves as the primary encoder for visual and linguistic inputs. It features a deep transformer architecture with 28 layers, a hidden dimension of 3584 and 28 attention heads. The vision encoder handles variable-resolution images and video frames, processing them into a sequence of visual embeddings.
    
    \item State Transformer (SALR Module): To align with the VLM backbone for layer-wise fusion, the State Transformer is designed with a matching depth of 28 layers. However, to maintain efficiency, it uses a smaller hidden dimension of 1024 and 8 attention heads. This component is responsible for processing the robot's proprioceptive state and performing the State-Aware Latent Re-representation at each layer.
    
    \item Action Expert: The downstream action generator is implemented as a Conditional Flow Matching (CFM) transformer decoder. It consists of 28 transformer layers with a hidden dimension of 1280 and 10 attention heads. This decoder takes the grounded representations from the SALR module as conditions to iteratively denoising the action sequence.
\end{itemize}

\subsection{Training Infrastructure and Parallelization}

Training a large VLA model on video sequences requires significant computational resources and memory optimization.

\textbf{Hyperparameters.}
We use the AdamW optimizer with $\beta_1=0.9, \beta_2=0.999$, and a weight decay of 0.01. The learning rate follows a cosine decay schedule with a warmup of 5,000 steps, peaking at $2.5 \times 10^{-5}$. The batch size is set to 1280. We train the model for approximately 14 days.

\textbf{Parallelization Strategy.}
We employ a cluster of 32 NVIDIA A100 (40GB) GPUs. To efficiently train the large model on video sequences while managing inter-node communication overhead, we adopt a Hybrid Sharding Strategy:

\begin{itemize}
    \item Intra-Node ZeRO-3: Within each node, we utilize Fully Sharded Data Parallel (FSDP) with ZeRO-3 optimization. This shards model parameters, gradients, and optimizer states across GPUs to maximize memory efficiency, fitting the large model into the limited 40GB VRAM.
    \item Inter-Node DDP: Across nodes, we employ standard Distributed Data Parallel (DDP). This reduces the communication frequency required for parameter gathering compared to global FSDP, thereby improving scaling efficiency on our cluster infrastructure.
    \item Gradient Checkpointing: We apply activation checkpointing across all transformer layers in the model, including the vision encoder, the VLM backbone, the State Transformer, and the Action Expert. This trades a small amount of compute overhead for significant memory savings, enabling the training of long-context sequences.
\end{itemize}

\section{Additional Experiments: BusyBox Bi-Manual Aloha}
\label{sec:additional_exp}

Due to space constraints in the main paper, we present the detailed results of our generalization experiments on the Bi-Manual Aloha BusyBox benchmark in this section. Figure~\ref{fig:additional_tasks} (Bottom) visualizes three representative interaction scenarios.

\subsection{Experimental Setup}
The BusyBox setup is designed to test a robot's ability to manipulate diverse, reconfigurable physical interfaces. It consists of six modular components: buttons, switches,  sliders, wires, a knob and a display.
\begin{itemize}
    \item \textbf{Task Definition:} We evaluate on 6 of all the canonical tasks defined in the BusyBox dataset, which include actions like ``turn the knob'', ``push the button'', ``move the slider'' and ``flip the switch''.
    \item \textbf{Metric:} We report the average Success Rate (SR) across all 6 tested tasks. A task is considered successful if the object state changes to the target state (e.g., the button pushed, the switch flied to the goal position) within a fixed time limit.
\end{itemize}

\subsection{Results and Analysis}
We compare LoLA against two baselines: Diffusion Policy \cite{chi2023diffusion} and $\pi_0$ \cite{black2410pi0}. All models were trained on the same subset of the BusyBox dataset.

Table~\ref{tab:busybox_results} summarizes the quantitative results. The results indicate that LoLA achieves competitive performance on this fine-grained bimanual manipulation benchmark. The BusyBox tasks require precise hand-eye coordination and the ability to generalize to different spatial configurations of the modules.
We attribute LoLA's performance gain (+16.7\% over $\pi_0$) to the State-Aware Latent Re-representation (SALR) module. By explicitly grounding the high-dimensional visual features of the small, intricate BusyBox components (like the small switches or wire tips) into the robot's physical end-effector space, LoLA can generate more precise action sequence, improving success in insertion and switching tasks.

\end{document}